\documentclass[letterpaper, 10 pt, conference]{ieeeconf}  

\IEEEoverridecommandlockouts                              

\usepackage{amsmath} 
\usepackage{amssymb}  
\usepackage{adjustbox}
\usepackage{bm}  
\usepackage{graphics} 
\usepackage{epsfig} 
\usepackage{wrapfig}
\usepackage{subcaption}
\usepackage[font=small,labelfont=small]{caption}
\usepackage{hyperref}
\usepackage{color}
\usepackage[table]{xcolor}
\usepackage[ruled, linesnumbered, vlined]{algorithm2e}
\usepackage{sidecap}
\usepackage{tikz}
\usetikzlibrary{arrows.meta}
\usepackage[noadjust]{cite}  
\usepackage{cuted}
\usepackage{multirow}
\usepackage{booktabs}
\usepackage{amsthm}  
\usepackage{soul}

\newcommand{\ty}[1]{{\scriptscriptstyle{\mathcal{#1}}}}

\newcommand{\euclideanspace}{\mathbb{R}}
\newcommand{\manifold}{\mathcal{M}}
\newcommand{\tangentspace}[1]{\mathcal{T}_{#1}\mathcal{M}}

\newcommand{\norm}[2]{\| #2\|_{#1}}  
\newcommand{\manifolddist}[2]{d_{\manifold}(#1, #2)}

\newcommand{\expmapblank}[1]{\text{Exp}_{#1}}  
\newcommand{\logmapblank}[1]{\text{Log}_{#1}}  
\newcommand{\expmap}[2]{\expmapblank{#1}\left(#2\right)}  
\newcommand{\logmap}[2]{\logmapblank{#1}\left(#2\right)}  
\newcommand{\prltrspblank}[2]{\Gamma_{#1 \rightarrow #2}}  
\newcommand{\prltrsp}[3]{\prltrspblank{#1}{#2}(#3)}  
\newcommand{\manifoldsqrdist}[2]{d_{\manifold}(#1, #2)^2}
\newcommand{\sphere}[1]{\mathcal{S}^{#1}}
\newcommand{\spd}[1]{\mathcal{S}_{\ty{++}}^{#1}}

\newcommand{\curve}{\bm{\gamma}}
\DeclareMathOperator*{\argmin}{argmin} 

\newtheorem{fal}{Fallacy}

\graphicspath{{Figures/}}

\title{\LARGE \bf
Unraveling the Single Tangent Space Fallacy: An Analysis and Clarification for Applying Riemannian Geometry in Robot Learning}

\author{No\'emie Jaquier$^{*1}$, Leonel Rozo$^{*2}$, and Tamim Asfour$^{1}$
\thanks{This work was supported by the Carl Zeiss Foundation under the project JuBot and the European Union's Horizon Europe Framework Programme under grant agreement No 101070596 (euROBIN).	}
\thanks{*These authors contributed equally (listed in alphabetical order).}
\thanks{$^{1}$Institute for Anthropomatics and Robotics, Karlsruhe Institute of Technology, Karlsruhe, Germany.
        \href{mailto:noemie.jaquier@kit.edu}{\textrm{noemie.jaquier@kit.edu}}} 
\thanks{$^{2}$Bosch Center for Artificial Intelligence. Renningen, Germany.}
}

\begin{document}

\makeatletter
\let\@oldmaketitle\@maketitle
\renewcommand{\@maketitle}{\@oldmaketitle
	\vspace{-13.5ex}
}
\makeatother
\maketitle

\thispagestyle{empty}
\pagestyle{empty}

\begin{strip}
    \centering
    \captionsetup{type=figure}
	\begin{subfigure}[b]{.2\linewidth}
		\includegraphics[trim={6cm 4cm 5.5cm 3cm}, clip,width=\textwidth]{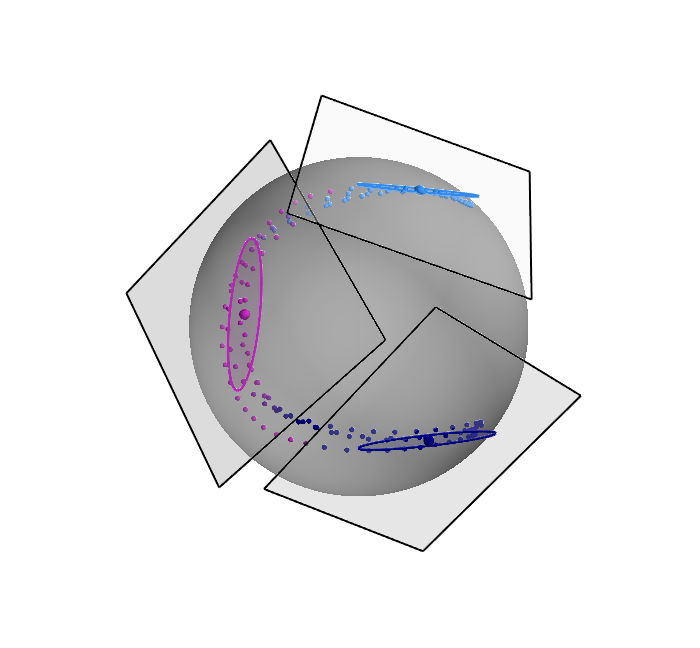}
		\includegraphics[trim={4cm 3cm 4cm 3cm}, clip,width=\textwidth]{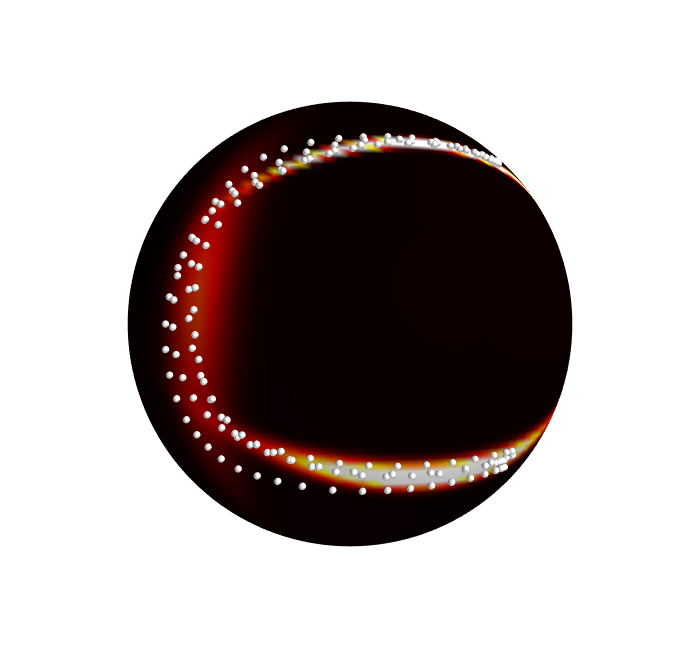}
		\caption{Riemannian } 
		\label{subFig:SphereRiemannianGMM}
	\end{subfigure}
 \hspace{0.3cm}
	\begin{subfigure}[b]{.2\linewidth}
		\includegraphics[trim={4cm 2cm 4cm 2cm}, clip,width=\textwidth]{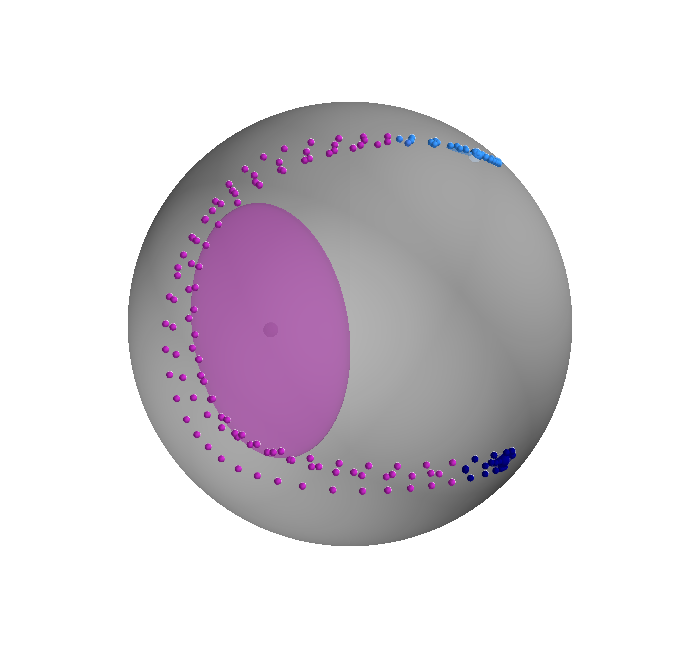}
		\includegraphics[trim={4cm 3cm 4cm 3cm}, clip,width=\textwidth]{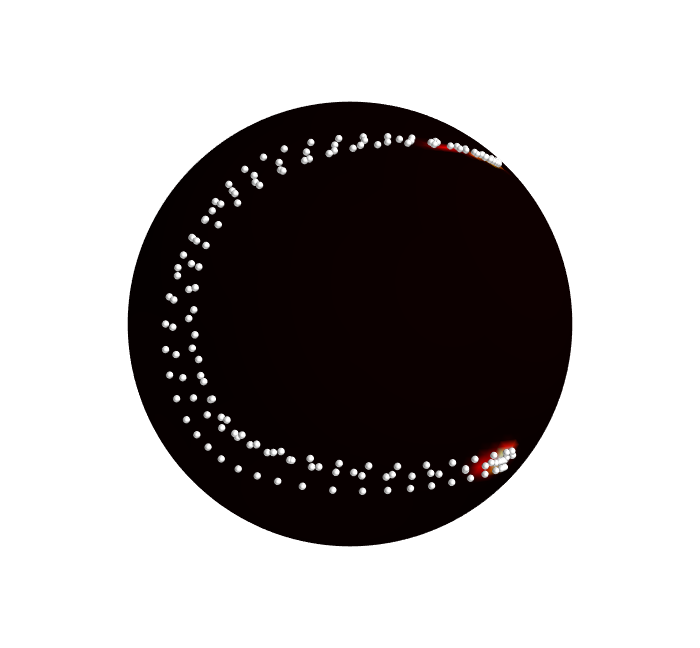}
		\caption{Euclidean } 
		\label{subFig:SphereEuclideanGMM}
	\end{subfigure}
  \hspace{0.3cm}
	\begin{subfigure}[b]{.2\linewidth}
		\includegraphics[trim={4cm 2cm 4cm 2cm}, clip,width=\textwidth]{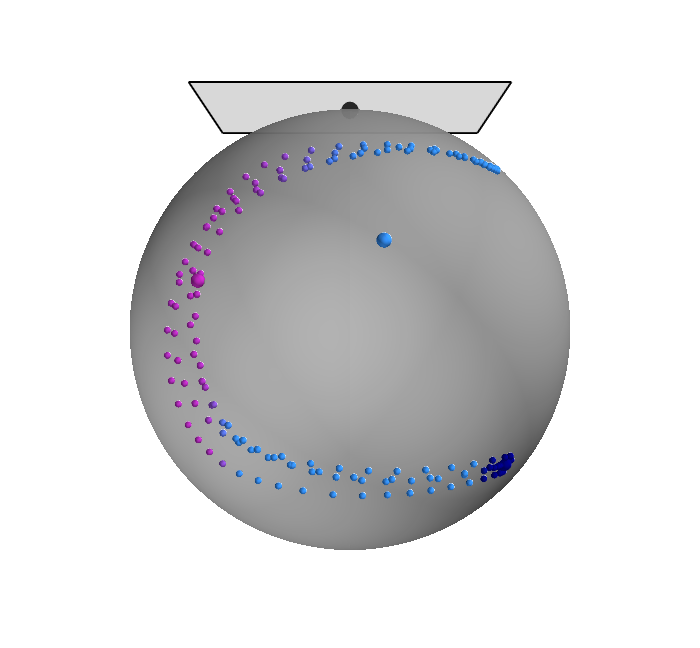}
		\includegraphics[trim={4cm 3cm 4cm 3cm}, clip,width=\textwidth]{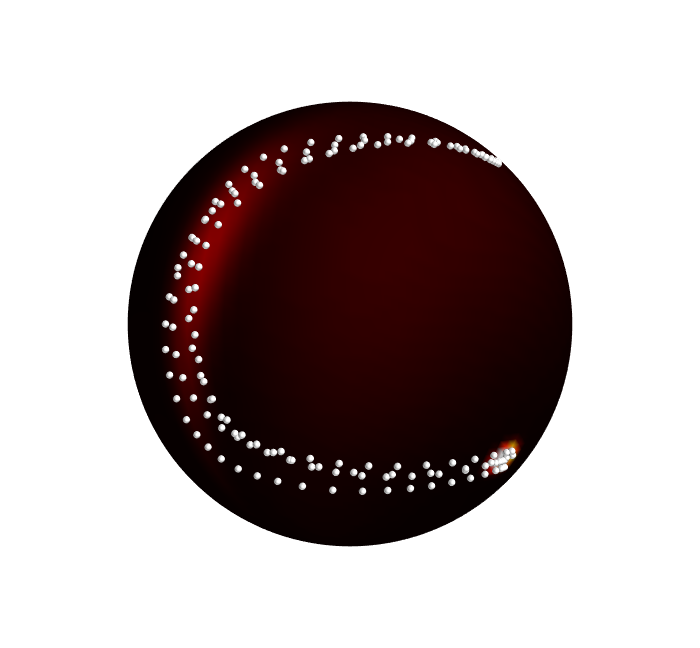}
		\caption{Tangent GMM (origin)} 
		\label{subFig:SphereNaiveGMM}
	\end{subfigure}
  \hspace{0.3cm}
	\begin{subfigure}[b]{.2\linewidth}
		\includegraphics[trim={4cm 2cm 4cm 2cm}, clip,width=\textwidth]{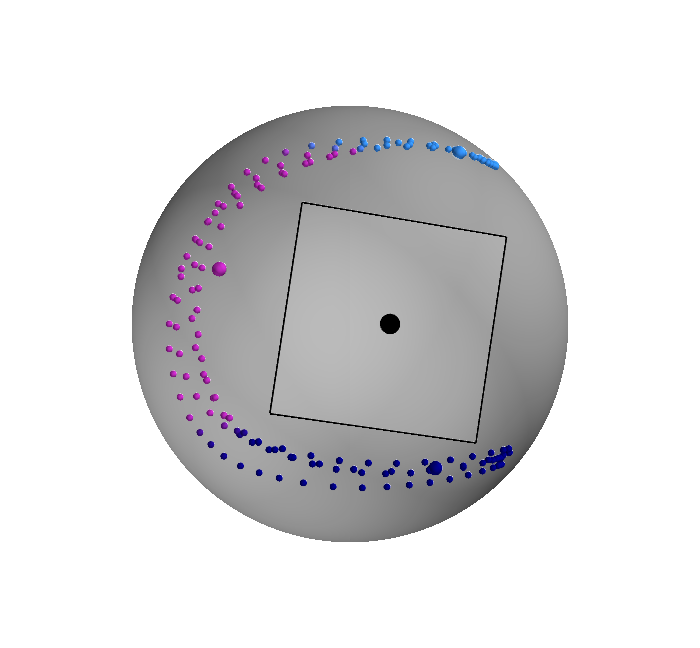}
		\includegraphics[trim={4cm 3cm 4cm 3cm}, clip,width=\textwidth]{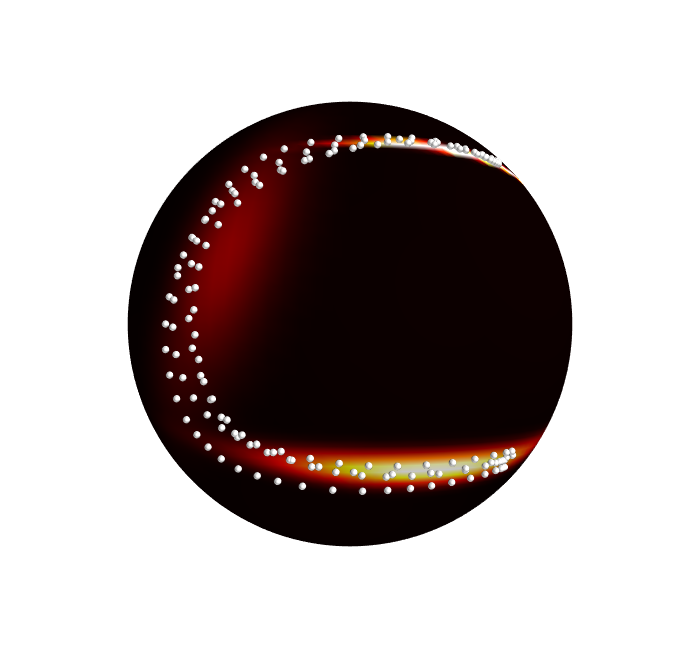}
		\caption{Tangent GMM (mean) } 
		\label{subFig:SphereNaiveMeanGMM}
	\end{subfigure}
	\caption{Illustration on the sphere $\sphere{2}$ of the importance of geometry in estimating GMM parameters and densities on Riemannian manifolds. \emph{Top row:} Estimated GMM parameters. The colors of the datapoints indicate their (shared) assignments to different GMM components. Two configurations of the single tangent space used by Tangent GMM are displayed in (c)-(d). \emph{Bottom row:} Estimated density on the sphere. Colors ranging from black to yellow indicate low to high probability density $p(\bm{x})$. Datapoints are depicted as white dots.}
	\label{Fig:Sphere_GMM_comparison}
 \vspace{-0.4cm}
\end{strip}

\begin{abstract}
In the realm of robotics, numerous downstream robotics tasks leverage machine learning methods for processing, modeling, or synthesizing data.
Often, this data comprises variables that inherently carry geometric constraints, such as the unit-norm condition of quaternions representing rigid-body orientations or the positive definiteness of stiffness and manipulability ellipsoids.
Handling such geometric constraints effectively requires the incorporation of tools from differential geometry into the formulation of machine learning methods.
In this context, Riemannian manifolds emerge as a powerful mathematical framework to handle such geometric constraints.
Nevertheless, their recent adoption in robot learning has been largely characterized by a mathematically-flawed simplification, hereinafter referred to as the ``single tangent space fallacy".
This approach involves merely projecting the data of interest onto a single tangent (Euclidean) space, over which an off-the-shelf learning algorithm is applied. 
This paper provides a theoretical elucidation of various misconceptions surrounding this approach and offers experimental evidence of its shortcomings.
Finally, it presents valuable insights to promote best practices when employing Riemannian geometry within robot learning applications.
\end{abstract}

\section{INTRODUCTION}
\vspace{-0.2cm}
The recent integration of machine learning techniques in robotics, spanning control, perception, and reasoning domains, has substantially enhanced robots' capabilities across diverse environments.
Nevertheless, a notable oversight in many of these approaches pertains to the intrinsic non-Euclidean geometry often inherent in robotic data. 
For instance, three-dimensional rotations can be viewed as elements of the special orthogonal group $\operatorname{SO}(3)$ or the sphere $\sphere{3}$, control gains, inertia, and manipulability ellipsoids lie in the space of symmetric-positive-definite (SPD) matrices~$\spd{d}$, the joint configuration of a $d$-degree-of-freedom robot with revolute joints can be seen as a point on a torus $\mathbb{T}^d$, while tree-like data can naturally be embedded in a hyperbolic space $\mathbb{H}^d$.
Under these conditions, Riemannian manifolds emerge as powerful mathematical structures used to model such curved spaces, allowing for a more accurate understanding of complex phenomena in various robot learning applications.
To perform computations on these manifolds, tangent spaces play a crucial role, as they provide local linear approximations of the manifold's geometry, enabling calculations of derivatives and vector operations.
However, the conventional approach of using a single tangent space for the entire manifold presents significant limitations that can lead to erroneous results and misinterpretations.

As proposed in several works~\cite{Fletcher04:TangentPCA,Tuzel08:RiemannianClassification,Abu-Dakka21:TangentKMPskills,Huang21:OrientationKMP}, the single tangent space approximation is carried out as follows: First, the Riemannian data or variable of interest is projected to a single tangent space; later, an off-the-shelf Euclidean learning algorithm models or processes the projected data; and finally the results are projected back on the Riemannian manifold. 
This simplification, though common, is fundamentally flawed and overlooks essential geometric properties of Riemannian manifolds.
Although few works have recently discussed the drawbacks of employing naive Euclidean approximations in different Riemannian settings~\cite{mathieu2020:RCNormalizingFlows,Jaquier21:IJRRManipulability,Zhang23:RiemannianStableDS}, a more theoretical and experimental analysis is still lacking in the current literature.
This paper is aimed at filling this gap by delving into the fallacies of employing a single tangent space for computations on Riemannian manifolds, and by providing experimental evidence of the inability of such an approach to capture the manifold's intrinsic curvature and its global structure, which can severely affect robot learning applications.

With the goal of rising awareness about the correct use of Riemannian methods in robot learning applications, \textbf{the main contributions of this paper are}: \emph{(1)} we present a simple yet elucidating example to introduce the concept and effects of the single tangent space fallacy; \emph{(2)} we point out and explain five common misconceptions of the single tangent space approach by building on several concepts of Riemannian geometry; and \emph{(3)} we experimentally illustrate the shortcomings of using a single tangent space in two common robot learning settings, namely, data density estimation and first-order dynamical systems learning.
Our experimental findings show that proper Riemannian formulations of a robot learning problem clearly outperform Euclidean and single-tangent space approximations.

\section{BACKGROUND}
\label{sec:Background}
\subsection{Riemannian Manifolds}
\begin{figure}
    \centering
    \begin{subfigure}[b]{.69\linewidth}
        \centering
        \includegraphics[trim={0.5cm 0.5cm 0.8cm 0.1cm}, clip,width=\linewidth]{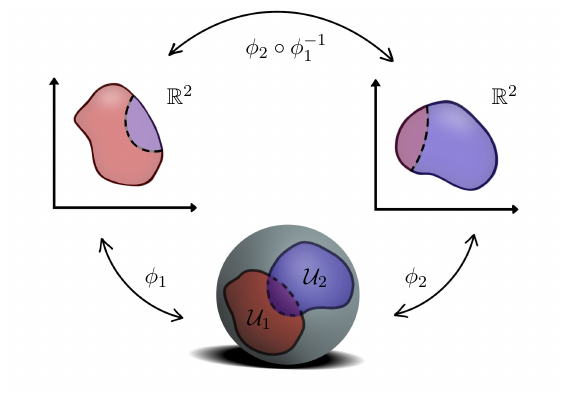}
        \label{fig:charts}
    \end{subfigure}
    \begin{subfigure}[b]{.29\linewidth}
    \centering
    \includegraphics[trim={2.0cm 0cm 1.0cm 0cm}, clip,width=\linewidth]{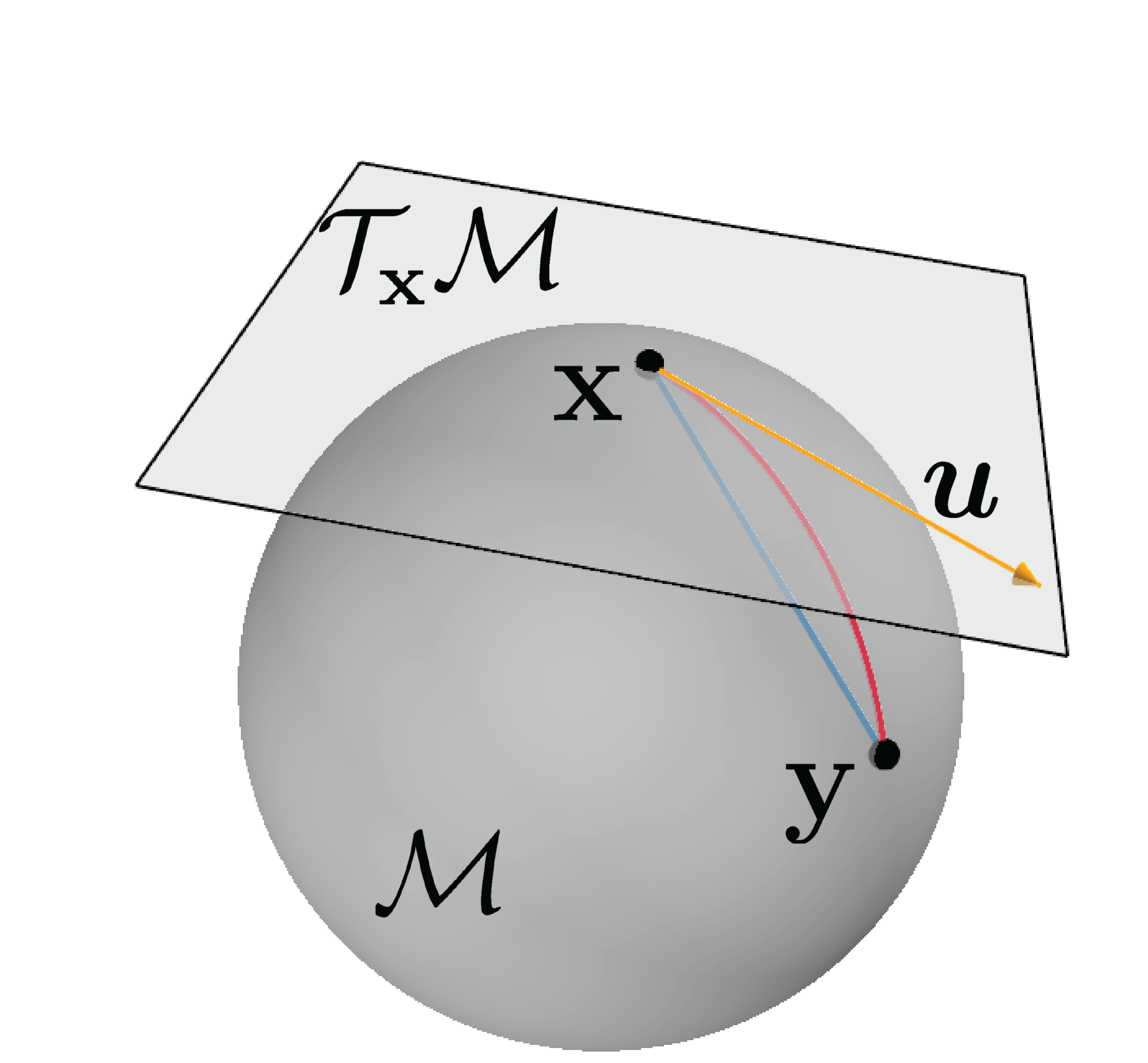}
    \includegraphics[trim={0cm 1.8cm 0.7cm 0cm}, clip,width=\linewidth]{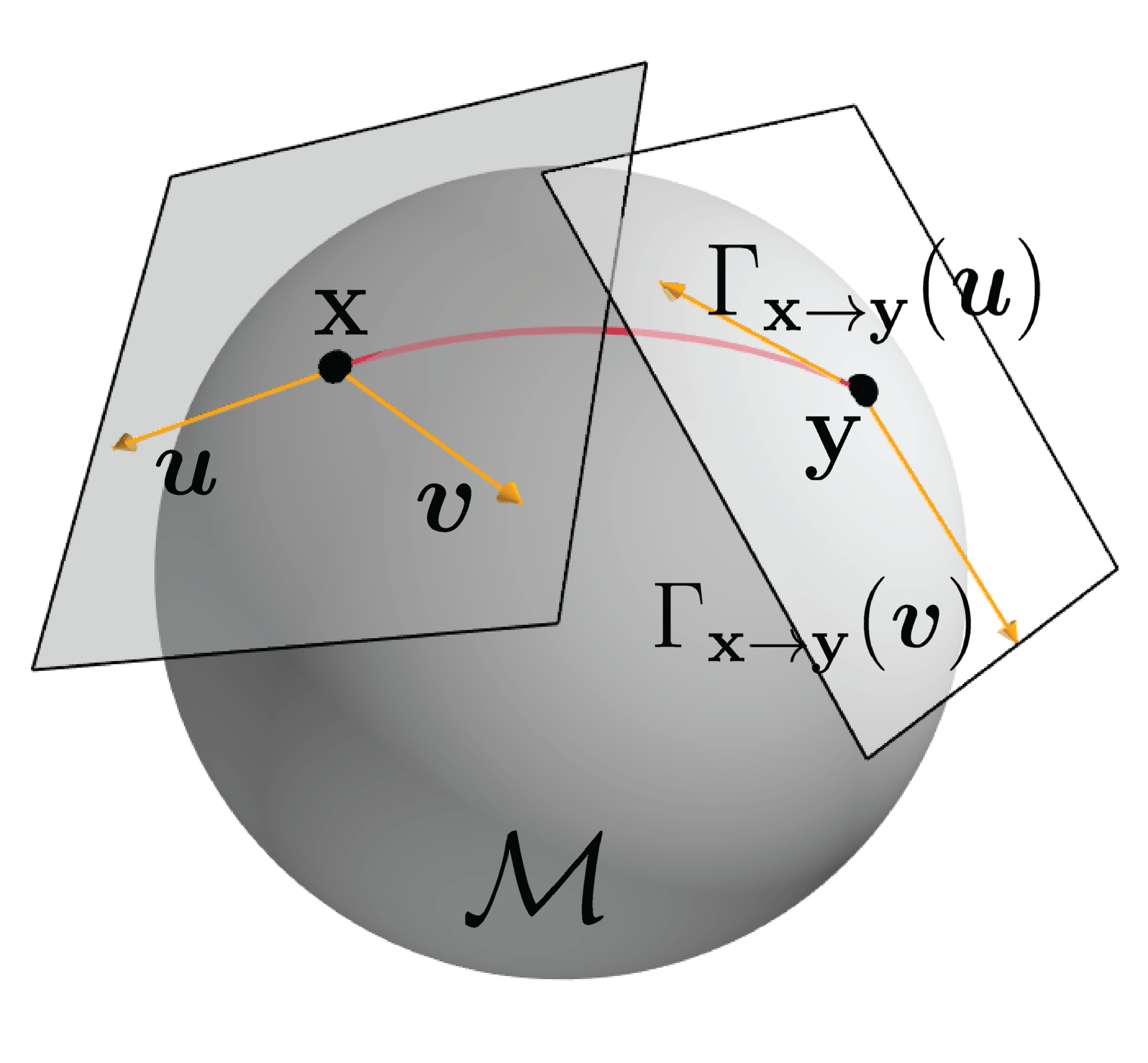}
    \label{fig:ExpLogMaps}
    \end{subfigure}
    \caption{Transition function between two charts on the sphere \emph{(left)}, and basic Riemannian operations \emph{(right)} on the sphere $\sphere{2}$. The vector $\bm{u}$ lies on the tangent space of $\bm{x}$ such that $\bm{u} = \text{Log}_{\bm{x}}(\bm{\mathrm{y}})$. $\Gamma_{\bm{\mathrm{x}}\rightarrow\bm{\mathrm{y}}}(\bm{u})$ and $\Gamma_{\bm{\mathrm{x}}\rightarrow\bm{\mathrm{y}}}(\bm{v})$ are parallel-transported vectors.}
    \label{fig:RiemannianBasics}
\end{figure}

A smooth manifold $\manifold$ can be intuitively conceptualized as a set of points that locally, but not globally, resemble the Euclidean space $\euclideanspace^d$~\cite{DoCarmo92:RiemannianGeometry, Lee18:RiemannianManifolds}. 
An abstract definition of a manifold encompasses the topological, differential, and geometric structure via the so-called \emph{charts}: maps between parts of $\manifold$ to $\euclideanspace^d$.
The collection of these charts --- seen as local parameterizations --- is called \emph{atlas}.
A chart on a smooth manifold $\manifold$ is a diffeomorphic mapping 
$\varphi: \mathcal{U} \to \tilde{\mathcal{U}}$ from an open set $\mathcal{U} \subset \manifold$ to an open set $\tilde{\mathcal{U}} \subseteq \euclideanspace^d$ (see Fig.~\ref{fig:RiemannianBasics}-\emph{left}).
The transition map between two intersecting sets $\mathcal{U}_1$ and $\mathcal{U}_2$, given by $\varphi_1 \circ \varphi_2^{-1}$ or $\varphi_2 \circ \varphi_1^{-1}: \mathbb{R}^d \rightarrow \mathbb{R}^d$ is also a diffeomorphism and guarantees consistency where the two charts overlap.
The smooth structure of $\manifold$ makes it possible to take derivatives of curves on the manifold, leading to tangent vectors in $\euclideanspace^d$.
The set of tangent vectors of all curves at $\bm{x} \in \manifold$ forms a $d$-dimensional affine subspace of $\euclideanspace^d$, known as the \emph{tangent space} $\tangentspace{\bm{x}}$ of $\manifold$ at $\bm{x}$.
The collection of all such tangent spaces is called the \emph{tangent bundle} $\tangentspace{}$. 
Finally, note that the local coordinates on $\mathcal{U}$ induced by the chart $\varphi$ define a basis for $\tangentspace{\bm{x}}$ (see Fig.~\ref{fig:RiemannianBasics}). 

A \emph{Riemannian manifold} $(\manifold, g)$ is a smooth manifold endowed with a Riemannian metric $g$, that is a family of inner products $g_{\bm{x}}: \tangentspace{\bm{x}} \times \tangentspace{\bm{x}} \rightarrow \euclideanspace$ associated to each point $\bm{x} \in \manifold$~\cite{Lee18:RiemannianManifolds}.
To operate with Riemannian manifolds, it is common practice to exploit their Euclidean tangent spaces and resort to mappings back and forth between $\tangentspace{\bm{x}}$ and $\manifold$. 
Specifically, the exponential map $\expmap{\bm{x}}{\bm{u}}: \tangentspace{\bm{x}} \to \manifold$ maps a point $\bm{u}\in\tangentspace{\bm{x}}$ to a point $\bm{y}$ on the manifold, so that it lies on the geodesic starting at $\bm{x}$ in the direction $\bm{u}$ and such that the geodesic distance $d_{\manifold}(\bm{x}, \bm{y}) = d_{\euclideanspace,g_{\bm{x}}}(\bm{x}, \bm{u})$ (see Fig.~\ref{fig:RiemannianBasics}-\emph{right}). 
The inverse operation is the logarithmic map $\logmap{\bm{x}}{\bm{y}}: \manifold \to \tangentspace{\bm{x}}$ .
Finally, the parallel transport $\prltrsp{\bm{x}}{\bm{y}}{\bm{u}}: \tangentspace{\bm{x}}\to\tangentspace{\bm{y}}$ describes how elements of $\manifold$ can be transported along curves on $\manifold$ while maintaining their intrinsic geometric properties. 
This allows us to operate elements lying on different tangent spaces (see Fig.~\ref{fig:RiemannianBasics}-\emph{right}).

\subsection{Riemannian Gaussian Distributions}
A Riemannian Gaussian distribution (RGD) $\mathcal{N}_{\manifold}(\bm{x} | \bm{\mu}, \bm{\Sigma})$, depends on two parameters, the mean $\bm{\mu} \in \mathcal{M}$ and covariance $\bm{\Sigma} \in \mathcal{T}_{\bm{\mu}}\mathcal{M}$. 
Its probability density function, generalising that of a Gaussian distribution on the Euclidean space $\euclideanspace^d$, can be locally approximated by\footnote{Under the assumption of having low-variance data.},
\begin{equation}
	\mathcal{N}_{\manifold}(\bm{x} | \bm{\mu}, \bm{\Sigma}) = \frac{1}{\sqrt{(2\pi)^d |\bm{\Sigma}|}} e^{-\frac{1}{2} \logmap{\bm{\mu}}{\bm{x}} \bm{\Sigma}^{-1} \logmap{\bm{\mu}}{\bm{x}}^{\top}} .
    \label{eq:RiemannianGaussian}
\end{equation}
This RGD corresponds to an approximated maximum-entropy distribution for Riemannian manifolds, as introduced in~\cite{Pennec06:BasicStatistics}.
The parameters of the RGD can be estimated via maximum likelihood estimation (MLE).
The MLE of $\bm{\mu}$ corresponds to the Fréchet mean $\bm{\mu}\in\manifold$~\cite{Frechet48:ManifoldMean} given by,
\begin{equation}
    \bm{\mu} = \argmin_{\bm{\mu}\in\manifold} \frac{1}{N} \sum_{n=1}^{N} \manifoldsqrdist{\bm{\mu}}{\bm{x}_n} .
    \label{Eq:Frechet_mean_definition}
\end{equation}
The covariance $\bm{\Sigma}$ is estimated in $\tangentspace{\bm{\mu}}$ as,
\begin{equation}
    \bm{\Sigma} = \frac{1}{N-1}\sum_{n=1}^{N}
    \logmap{\bm{\mu}}{\bm{x}_n} \logmap{\bm{\mu}}{\bm{x}_n}^\top .
    \label{Eq:Riemannian_unbiased_covariance}
\end{equation}

\section{LEARNING RIEMANNIAN DATA DISTRIBUTIONS: A MOTIVATIONAL EXAMPLE}
\label{sec:MotivationalExample}
We begin our analysis by studying a simple density estimation problem on the sphere $\sphere{2}$. 
Modeling complex, multimodal probability distributions on smooth manifolds is at the heart of various robotics tasks ranging from state estimation~\cite{Forster17:StateEstimation} to motion planning~\cite{Park08:StateEstimationMotionPlanningLieGroups} and learning from demonstrations~\cite{Zeestraten17riemannian,Jaquier21:IJRRManipulability}.
We aim at estimating the underlying density of $\mathsf{C}$-shaped data lying on the sphere $\sphere{2}$ with a simple, yet popular probabilistic model, i.e., a Gaussian mixture model (GMM) with $K=3$ components.  
As our data is not Euclidean, several GMM variants can be considered for estimating their density. 
All are initialized identically.

\subsection{Euclidean GMM}
As $\sphere{2} \subset \euclideanspace^3$, a first straightforward approach is to apply a traditional Euclidean GMM acting on the embedding space $\euclideanspace^3$. The corresponding density is of the classical form,
\begin{equation}
    p(\bm{x}) = \sum_{k=1}^K \pi_k \mathcal{N}(\bm{x} | \bm{\mu}_k, \bm{\Sigma}_k),
\end{equation}
where the mean $\bm{\mu}_k$, covariance $\bm{\Sigma}_k$, and prior $\pi_k$ of each Gaussian component are estimated via MLE.
Fig.~\ref{subFig:SphereEuclideanGMM} displays the estimated model and resulting density on $\sphere{2}$. We observe that the GMM components are located inside the sphere, as they are not constrained to lie on its surface. 
By acting on $\euclideanspace^3$, the model fully disregards the intrinsic data geometry.
As a result, the estimated density is mostly concentrated inside the sphere and fails to represent the observed data.

\subsection{Euclidean GMM on a Single Tangent Space}
\begin{figure}[tbp]
	\centering
	\includegraphics[trim={6cm 2cm 6cm 7.5cm}, clip, width=.35\linewidth]{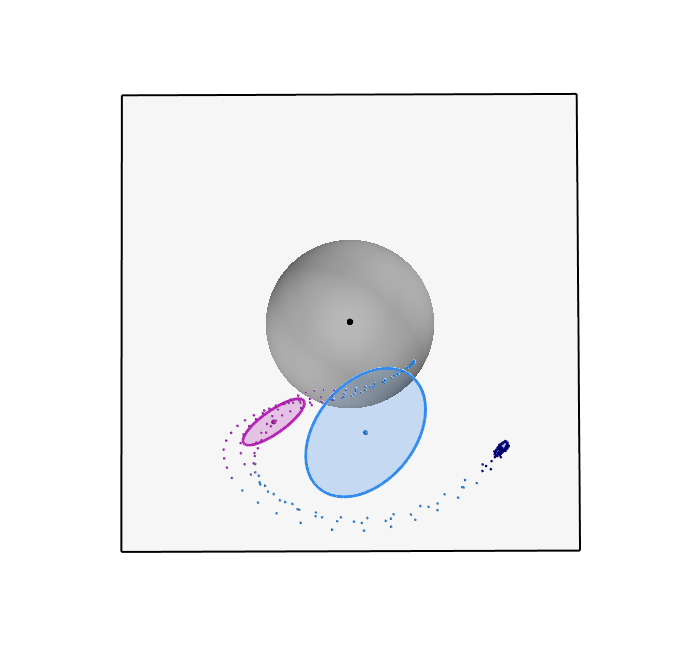}
    \hspace{0.5cm}
	\includegraphics[trim={4.5cm 3cm 5cm 3cm}, clip, width=.35\linewidth]{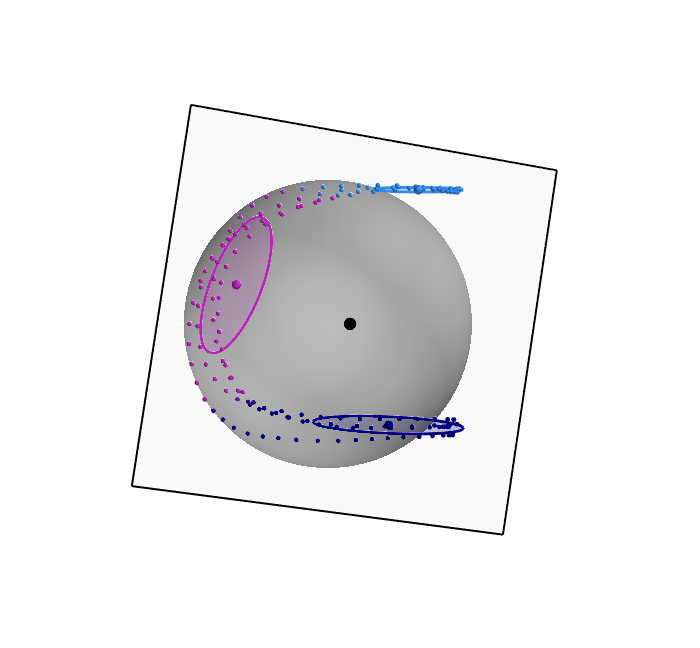}
	\caption{Datapoints and GMM of Figs.~\ref{subFig:SphereNaiveGMM},~\ref{subFig:SphereNaiveMeanGMM} in the single tangent spaces at the origin with a top view \emph{(left)} and  Fréchet mean \emph{(right)}.}
	\label{Fig:Sphere_GMM_comparison_tangent_spaces}
 \vspace{-0.6cm}
\end{figure}

Next, we apply a second naive approach, referred to as Tangent GMM, which consists in projecting all datapoints in a single tangent space using the logarithmic map, fitting a Euclidean GMM in this tangent space, and projecting the resulting probability density back onto the manifold using the exponential map.  
This approach might seem appealing as it uses an off-the-shelf Euclidean GMM, and the resulting density lies on the sphere surface. 
However, it displays various shortcomings, as shown next.
Fig.~\ref{subFig:SphereNaiveGMM} shows the Tangent GMM estimated via the tangent space at the origin.
Despite the post-projection to $\sphere{2}$, the model poorly fits the data: The mean of the light-blue component is located far away from the data and the density inside the $\mathsf{C}$ is non-zero, although this region is out of the data support. 
This is due to the severe distortions induced by the projection onto the Euclidean tangent space 
As shown in Fig.~\ref{Fig:Sphere_GMM_comparison_tangent_spaces}-\emph{left}, the $\mathsf{C}$-shape horizontally elongates and its upper part is reversely bent.  

Importantly, such distortions depend on the choice of tangent space. 
Fig.~\ref{subFig:SphereNaiveMeanGMM} shows the Tangent GMM applied at the Fréchet mean of the data, i.e., via the tangent space resulting in minimal distortions (see Fig.~\ref{Fig:Sphere_GMM_comparison_tangent_spaces}-\emph{right}). 
However, the model still fails at entirely capturing the underlying data distribution: It estimates a close-to-zero density at the bottom-left part of the $\mathsf{C}$, and a high density around the mean of the pink component, located out of the data support. 

\subsection{Riemannian GMM}
\vspace{-0.1cm}
Last, we examine a Riemannian GMM~\cite{SimoSerra14:GeodesicMixtureModels}, whose probability density is based on multiple RGDs~\eqref{eq:RiemannianGaussian}, so that,
\begin{equation}
    p(\bm{x}) = \sum_{k=1}^K \pi_k \mathcal{N}_{\manifold}(\bm{x} | \bm{\mu}_k, \bm{\Sigma}_k).
\end{equation}
In other words, each mixture component is centered at a point $\bm{\mu}_k\in\manifold$, while its directional dispersion is characterized by a covariance $\bm{\Sigma}_k\in\tangentspace{\bm{\mu}_k}$. The GMM parameters are estimated via MLE with a Riemannian Expectation-Maximization algorithm~\cite{SimoSerra14:GeodesicMixtureModels}.
Fig.~\ref{subFig:SphereRiemannianGMM} depicts the resulting Riemannian GMM. The learned model is consistent with the data geometry, and the estimated density matches the underlying data distribution, achieving the highest likelihood among the considered GMMs (see $\mathsf{C}$ column in Table~\ref{tab:ResultsLLGMM-sphere}).
By simultaneously leveraging multiple tangent spaces, the Riemannian GMM limits the distortions of individual tangent spaces, while guaranteeing that the estimated probability density exclusively lies on the manifold. As explained next, considering multiple tangent spaces is key to the design of high-performance geometric learning models.

\section{THE SINGLE TANGENT SPACE FALLACY}
\vspace{-0.1cm}
\label{sec:Fallacy}
The example of Section~\ref{sec:MotivationalExample} illustrated some of the limitations of learning on a single tangent space. 
The flaws of this approach were evident even on a relatively simple learning setting on a low-dimensional sphere, i.e., the simplest Riemannian manifold to operate on after $\euclideanspace^d$.
Here, we delve into the mathematical aspects that explain why most single-tangent-space-based approaches are fundamentally flawed, and highlight the common misunderstandings leading to their usage. 
Section~\ref{sec:Experiments} then illustrates their flaws experimentally.

\subsection{A Deep Dive into Mathematics of Tangent Spaces}
\vspace{-0.1cm}
To understand the single tangent space fallacy, we first formally review the notions of tangent vectors and spaces. Here, we define tangent vectors via an equivalence class of curves as in~\cite[Chap. 8]{Boumal23:IntroOptSmoothManifolds},~\cite[Chap. 3]{Lee13:SmoothManifolds}. 
We consider the set of all smooth curves $\curve: I \to \manifold$, passing through the point $\bm{x}\in\manifold$ such that $\curve(0)=\bm{x}$, with $I$ an interval in $\euclideanspace$. 
Given a chart $(\mathcal{U}, \varphi)$ at $\bm{x}$, two curves $\curve_1$ and $\curve_2$ are equivalent at $\bm{x}$ if and only if their derivatives equal at $\bm{x}$, i.e.,
\begin{equation}
    \curve_1 \sim \curve_2 \iff (\varphi\circ\curve_1)'(0)=(\varphi\circ\curve_2)'(0).
\end{equation}
The equivalence class of such curves, denoted $[\curve] = \curve'(0)$, defines a \emph{tangent vector}.  
The \emph{tangent space} $\tangentspace{\bm{x}}$ is then the set $\{[\curve]\}$ of all tangent vectors at $\bm{x}$.
The foregoing definitions lead to crucial insights regarding the single tangent space fallacy. 
First, tangent vectors are \emph{attached} to a specific point $\bm{x}$. 
Second, a \emph{linear structure} is naturally induced on $\tangentspace{\bm{x}}$ via the bijective map $d\varphi_{\bm{x}}: \tangentspace{\bm{x}} \to \euclideanspace^d: [\curve] \mapsto (\varphi\circ\curve)'(0)$,
as it satisfies, independently of the choice of chart,
\begin{equation}
    a [\curve_1] + b [\curve_2] = d\varphi_{\bm{x}}^{-1} \left( a\;\! d\varphi_{\bm{x}}([\curve_1]) + b \;\!d \varphi_{\bm{x}}([\curve_2]) \right).
\end{equation}
In other words, two tangent vectors can be added into a third, and a real multiple of a tangent vector remains a tangent vector.
Third, each tangent space $\tangentspace{\bm{x}}$ is a \emph{local linear approximation} of $\manifold$ at $\bm{x}$. 
As such, tangent spaces do not globally capture the intrinsic manifold geometry.

\begin{table*}[t!]
    \setlength{\tabcolsep}{2pt}
    \renewcommand{\arraystretch}{1.05}
    \centering
    \begin{tabular}{c|c|ccc|ccc|ccc|}
        \multicolumn{1}{c|}{} & \multicolumn{1}{c|}{$\mathsf{C}$} &\multicolumn{3}{c|}{Mixture of RGDs} &\multicolumn{3}{c|}{Mixture of WGDs} &\multicolumn{3}{c|}{Mixture of vMFs} \\
         $d$ & 2 & 3&7&10& 3&7&10& 3&7&10 \\
         \hline
        Euclidean GMM &$552$ & $-132\pm49$ & $-299\pm69$ & $-425\pm70$ & $-74\pm151$ & $-210\pm249$ & $-300\pm250$ & $195\pm68$ & $240\pm92$ & $241\pm122$\\
		Tangent GMM &$2117$ & $814\pm38$ & $514\pm66$ & $232\pm82$ & $809\pm103$ & $470\pm299$ & $284\pm254$ & $1010\pm44$ & $1010\pm77$ & $949\pm122$\\
        Riemannian GMM &$\bm{2176}$ & $\bm{853\pm40}$ & $\bm{642\pm66}$ & $\bm{432\pm92}$ & $\bm{850\pm98}$ & $\bm{543\pm291}$ & $\bm{318\pm334}$ & $\bm{1052\pm40}$& $\bm{1080\pm104}$& $\bm{1068\pm156}$\\
	    \hline
 \end{tabular}
     \caption{Average test log-likelihood over $100$ target densities estimated via GMMs on sphere manifolds $\sphere{d}$. }
    \label{tab:ResultsLLGMM-sphere}
    \vspace{-0.6cm}
\end{table*}

\begin{table}[t!]
    \setlength{\tabcolsep}{2pt}
    \renewcommand{\arraystretch}{1.05}
    \centering
    \begin{tabular}{c|cc|cc|}
        \multicolumn{1}{c|}{} &\multicolumn{2}{c|}{Mixture of RGDs} &\multicolumn{2}{c|}{Mixture of inv. Wisharts} \\
         $d$ & 2&3& 2&3 \\
         \hline
        Eucl. GMM & $-552\pm143$ & $-1228\pm218$ & $-381\pm133$ & $-860\pm218$\\
		Tan. GMM& $-409\pm55$ & $-909\pm112$ & $-207\pm89$ & $-355\pm158$ \\
        Riem. GMM& $\bm{-380\pm48}$ & $\bm{-792\pm136}$ & $\bm{-140\pm83}$ & $\bm{-134\pm133}$ \\
	    \hline
 \end{tabular}
    \caption{Average test log-likelihood over $100$ target densities estimated via GMMs on SPD manifolds $\spd{d}$.}
    \label{tab:ResultsLLGMM-spd}
    \vspace{-0.6cm}
\end{table}

\subsection{Fallacies of Single Tangent Space Approaches}
\vspace{-0.1cm}
Single-tangent-space-based approaches might appear appealing as \emph{(1)} the linear structure of tangent spaces complies with traditional Euclidean methods, and \emph{(2)} the results obtained on a tangent space can be mapped onto the manifold via the exponential maps. 
However, such approaches are fundamentally flawed. 
Next, we explain five core misconceptions, which relate to the single tangent space fallacy.

\begin{figure}
    \centering
    \includegraphics[trim={1.1cm 2.2cm 2.2cm 3.2cm}, clip,width=.36\linewidth]{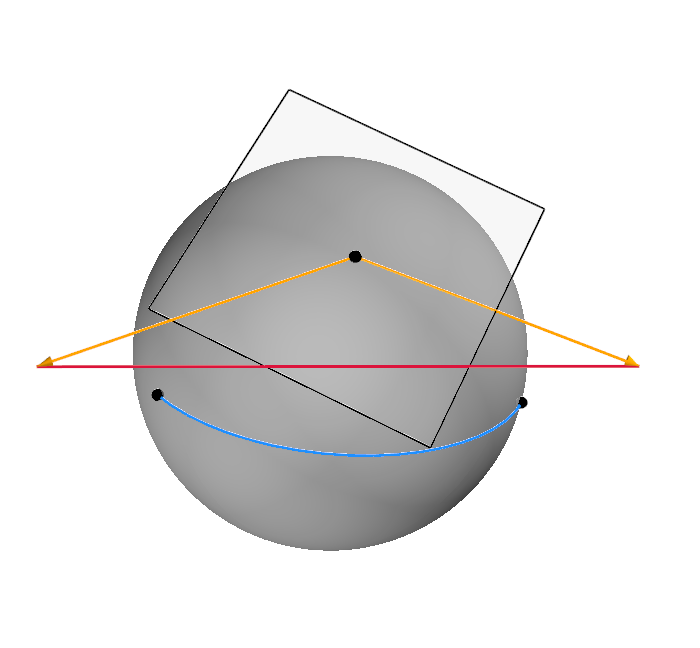}
    \includegraphics[trim={5cm 1.0cm 5.cm 6cm}, clip,width=.3\linewidth]{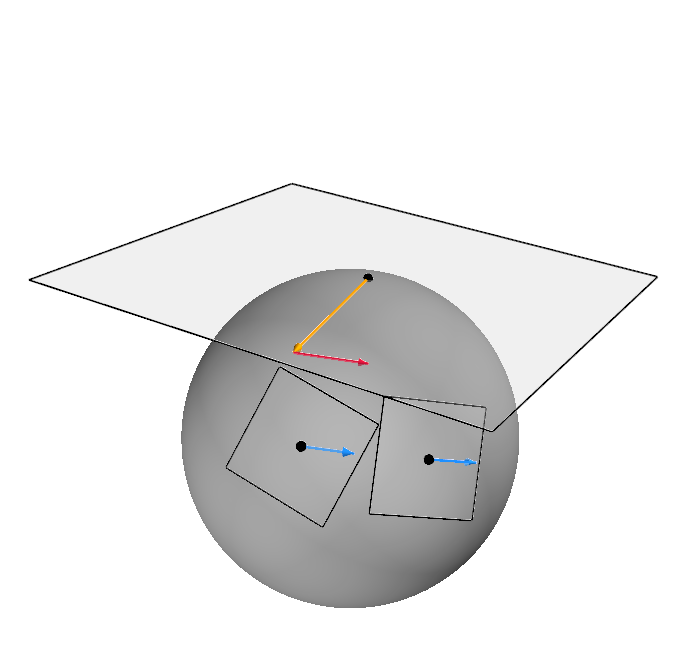}
    \includegraphics[trim={4.0cm 5.2cm 10cm 6.5cm}, clip,width=.3\linewidth]{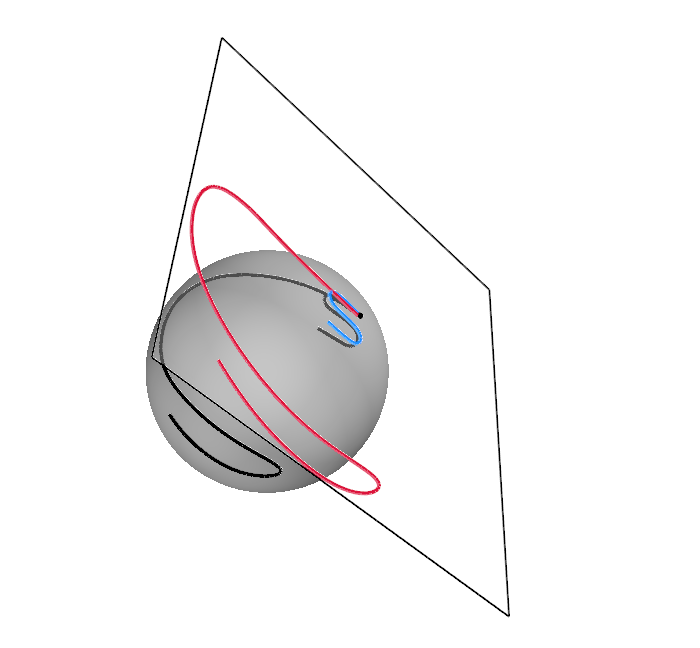}
    \caption{Illustration of the fallacies $2$, $3$, and $5$ (in red). \emph{Left:} the blue and red curves depict the geodesic and Euclidean distances, respectively. \emph{Middle:} the blue arrows represent two velocities w.r.t two different tangent spaces, while the red arrow depicts a linear velocity on the single tangent space. \emph{Right:} the black curves display Riemannian trajectories, while the blue and red curves show the corresponding distorted projections on a single tangent space.}
    \label{fig:fallacies}
    \vspace{-0.5cm}
\end{figure}

As mentioned earlier, exponential and logarithmic maps are key operations to leverage the linear structure of tangent spaces. However, two key aspects of these mappings are often overlooked when operating on a single tangent space.
\begin{fal}
    The exponential map is a local diffeomorphism.
\end{fal}
Namely, it is a diffeomorphism between a neighborhood $\mathcal{B}_r(\bm{0})\subseteq\tangentspace{\bm{x}}$ around $\bm{0}\in\tangentspace{\bm{x}}$ and the manifold $\manifold$. 
The neighborhood $\mathcal{B}_r(\bm{0})$ is characterized by the injectivity radius $r$, so that it neither contains nor goes beyond the cut locus of $\manifold$.
In other words, the injectivity radius fundamentally quantifies the extent to which geodesics behave like straight lines in $\euclideanspace^d$ within the largest possible neighborhood $\mathcal{B}_r$.
For instance, the sphere exponential map is diffeomorphic only for tangent vectors $\bm{u}$ satisfying $\norm{\bm{x}}{\bm{u}}<d_{\sphere{d}}(\bm{x}, \bm{-x})=r$. 
Off-the-shelf Euclidean learning approaches applied on a single tangent space disregard the injectivity radius. 
Thus, they are prone to generate different tangent vectors associated to the same point on the manifold, leading to inconsistencies.

\begin{fal}
    In general, the exponential map is not an isometric mapping and consequently does not preserve distances.
\end{fal}
As local approximations, tangent spaces intrinsically distort distances as, e.g., in Figs.~\ref{Fig:Sphere_GMM_comparison_tangent_spaces},~\ref{subFig:SphereNaiveMeanDS_TangentSpace}. 
The following inequality holds for manifolds with non-zero curvature $\forall \;  \bm{y}_1, \bm{y}_2 \neq \bm{x}$,
\begin{equation}
\manifolddist{\bm{y}_1}{\bm{y}_2} \neq \norm{\bm{x}}{\logmap{\bm{x}}{\bm{y}_2} - \logmap{\bm{x}}{\bm{y}_1}},
\end{equation}
so that only distances with respect to the basepoint $\bm{x}$ are preserved (see Fig.~\ref{fig:fallacies}-\emph{left}). 
In particular, distances are under/over-estimated on tangent spaces of manifolds with constant positive/negative curvature.
Therefore, applying distance-based learning algorithms on data projected in a single tangent space unavoidably leads to distorted models as they overlook the intrinsic geometry of the manifold. 
Distortions are more pronounced for data lying far away from the basepoint and exacerbated for complex Riemannian manifolds, e.g., non-constant curvature and high-dimensional manifolds. 
For instance, orientation kernelized movement primitives based on a single tangent space~\cite{Huang21:OrientationKMP} learn distorted models that can only be applied to data closely grouped on $\sphere{3}$. 
This limitation is further worsened on symmetric positive-definite matrix (SPD) manifolds~\cite{Abu-Dakka21:TangentKMPskills}.
Leveraging several tangent spaces significantly reduces distortions and increases the performance of learning algorithms, e.g., for clustering~\cite{SimoSerra17:GeodesicMixtureModels}, and classification~\cite{Sanin12:KtangentSpacesClassification} tasks, among others.

\begin{fal}
    Velocities on Riemannian manifolds are attached to specific positions and are not independent entities.
    \vspace{-0.1cm}
\end{fal}
This implies that Riemannian vector fields, e.g., first-order dynamical systems (DS), assign a tangent vector in $\tangentspace{\bm{x}}$ to each point $\bm{x}$ on the manifold. 
Such vectors are attached to $\bm{x}$ and associated with the physical notion of velocity. 
In contrast, vectors $\bm{z}=\logmap{\bm{x}}{\bm{y}_2} - \logmap{\bm{x}}{\bm{y}_1}$ resulting from tangent space projections with $\bm{y}_1, \bm{y}_2 \neq \bm{x}$ are physically-meaningless and can neither be associated with nor be integrated as velocities at $\bm{y}_1$ or $\bm{y}_2$  (see Fig.~\ref{fig:fallacies}-\emph{middle}). 
Hence, a DS learned on a single tangent space~\cite{Wang22:DeepRiemannianSkills} unavoidably leads to distortions and to physically-incoherent models. 
Riemannian second-order DS~\cite{Fiori22:RiemannianLie2DS} are intrinsically formulated via tangent vectors and covariant accelerations~\cite{RojasQuintero22:RiemannianPontryagin}, and cannot be learned in a single tangent space. 

\begin{fal}
    In general, data on Riemannian manifolds and Lie groups are not treated equivalently.
\end{fal}
Several Riemannian manifolds can be endowed with a Lie group structure. 
This allows the formulation of learning algorithms that leverage their Lie algebra, i.e., the tangent space at the group identity. 
Exponential and logarithmic maps on Lie groups are mappings between the Lie algebra and the group. 
These are often defined in combination with composition operations~\cite{Sola18:LieTheory}, which intuitively play the role of transportation operators w.r.t the group identity (a.k.a. the adjoint operator). 
This differs from the Riemannian exponential and logarithmic maps. 
Therefore, learning on the Lie algebra is not equivalent to learning on a single tangent space of a Riemannian manifold. 
Learning algorithms must be formulated adequately for each type of manifold, see, e.g., the differences between learning stable DS on Lie groups~\cite{Urain22:LieAlgebraDS} and on Riemannian manifolds~\cite{Zhang23:RiemannianStableDS}.

\begin{fal}
    Experiments on constant curvature and low-dimensional Riemannian manifolds may be misleading.
    \vspace{-0.1cm}
\end{fal}
Inaccuracies and distortions induced by using a single tangent space are less apparent under certain conditions. 
For instance, Sommer et al.~\cite{Sommer14:ExactPGA} showed that the variance of certain symmetric datasets on spheres are equivalently captured via Tangent and Exact Principal Geodesic Analysis (PGA)~\cite{Fletcher04:TangentPCA}, while significant differences arise for similar datasets on hyperbolic spaces. 
Moreover, the reconstruction error of Tangent PGA drops significantly faster than Exact PGA for increasing data variance~\cite{Fan22:NestedHyperbolic}.
Such behaviors lead to inaccurate and overly-optimistic interpretations of results obtained via single tangent space-based approaches on small data ranges  (see Fig.~\ref{fig:fallacies}-\emph{right}), as in~\cite{Abu-Dakka21:TangentKMPskills,Wang22:DeepRiemannianSkills}, and on low-dimensional simple manifolds, e.g., spheres as in~\cite{Huang21:OrientationKMP}. 

\subsection{Cases Where a Single Tangent Space Is Appropriate}
\vspace{-0.08cm}
Generalization of linear methods are often intrinsically formulated via a single tangent space. 
These include geodesic regression~\cite{Fletcher13:GeodesicRegression,Kim14:MGLM} and geodesic basis function models of movement primitives~\cite{Rozo22:OrientationProMPs}. 
Such learning models are based on a single geodesic, which is completely defined by a basepoint $\bm{x}\in\manifold$ and an initial velocity $\bm{v}\in\tangentspace{\bm{x}}$, i.e., via a single tangent space $\tangentspace{\bm{x}}$. 
Consequently, learning naturally occurs in $\tangentspace{\bm{x}}$ only. 
Notice that an appropriate basepoint $\bm{x}$ must still be selected to achieve optimal performance.

\section{EXPERIMENTS}
\vspace{-0.1cm}
\label{sec:Experiments}
We here illustrate the single tangent space fallacy via two different robot learning applications. 
\vspace{-0.05cm}

\subsection{Learning Densities}
\vspace{-0.1cm}
We return to the density estimation problem of Section~\ref{sec:MotivationalExample} and analyse the performance of Euclidean GMM, Tangent GMM at the Fréchet mean, and Riemannian GMM to estimate several target densities on various Riemannian manifolds.  
We train the models on $100$ samples from the target densities and compute their likelihood on $100$ additional test points. 
In each case, all GMMs are initialized identically. 

First, we estimate densities on spheres $\sphere{d}$ with $d=\{3, 7, 10\}$. 
We consider target densities in the form of mixtures of $3$ RGDs, $3$ Wrapped Gaussian distributions (WGDs)~\cite{Mardia99:DirectionalStats,Galaz-Garcia22:WrappedHomogenous}, or $3$ isotropic von Mises-Fisher distributions (vMFs)~\cite{Mardia99:DirectionalStats}. 
The means of the $3$ mixture components are defined as $\bm{\mu}_1= \frac{1}{d+1}\bm{1}_{d+1}$, 
$\bm{\mu}_2=\frac{1}{d+1}\left(\begin{smallmatrix}
    1 \\ -\bm{1}_{d}
\end{smallmatrix}\right)$, and
$\bm{\mu}_3=\frac{1}{d+1}\left(\begin{smallmatrix}
    -1 \\ \bm{1}_{d}
\end{smallmatrix}\right)$, with $\bm{1}_d$ denoting a size $d$ vector of ones. 
The covariances $\bm{\Sigma}_k$  
of the RGDs and WGDs are randomly-generated SPD matrices, and the concentration parameters of the vMFs are randomly generated from a uniform distribution on $[20, 70]$. 
As show in Table~\ref{tab:ResultsLLGMM-sphere}, the Riemannian GMMs consistently outperform the Euclidean and Tangent GMMs, thus matching best the target densities.  
Although Tangent GMMs largely outperform Euclidean GMMs, their performance remains below that of the Riemannian models. 
This performance gap increases for complex target densities, e.g., mixtures of RGDs and WGDs, compared to simpler densities defined by isotropic components (e.g., vMFs), which are easier to model on a single tangent space. 
Note that the performance gap also grows with the manifold dimensionality.

Next, we study manifolds with non-constant curvature, and estimate densities on SPD manifolds $\spd{d}$ of dimensions $d=\{2,3\}$. 
We consider target densities in the form of mixtures of $5$ RGDs, or $5$ inverse-Wishart distributions (iWDs)~\cite{Gelman13:Bayesian}. 
The mean and covariance of the RGDs are randomly generated as SPD matrices of appropriate dimensions with eigenvalues in $[0.1, 2]$ and $[0.1, 0.5]$, respectively. 
The scale matrix and degree of freedom of the iWDs are generated as the means of the RGD, and from a uniform distribution on $[d+1, d+3]$, respectively. 
The Riemannian GMM consistently achieves the highest likelihood and, similarly as for sphere manifolds, the performance gap between Riemannian GMMs and the other approaches grows with the manifold dimensionality.

\subsection{Learning Dynamical Systems with Normalizing Flows}
\label{subsec:ExperimentsDS}
\vspace{-0.1cm}
We shift our attention to the problem of learning stable first-order DS, whose state evolves on a Riemannian manifold. 
Specifically, we consider DS of the form $\Dot{\bm{x}} = f(\bm{x})$, with $\bm{x} \in \manifold$ and $\Dot{\bm{x}} \in \tangentspace{\bm{x}}$. 
Such systems are conventionally learned from demonstrations, by using example trajectories of the desired motion to train a parametrized function $f_{\bm{\theta}}$ under certain stability guarantees.
Most prior works assume the state $\bm{x}$ to evolve in an Euclidean space~\cite{KhansariZadeh2011:StableEstimatorDS,Neumann2014:SDS_diffeomorphism,Blocher17:LearningContractiveDS,Figueroa18:LearningPhysicsDS,Rana2020:EuclideanizingFlows,Zhang2022:accurateSDS}. 
However, there has been a growing interest in extending these approaches to non-Euclidean settings~\cite{Zhang23:RiemannianStableDS,Wang22:DeepRiemannianSkills,Urain22:LieAlgebraDS}.

\begin{figure*}
\vspace{-0.25cm}
        \centering
    \captionsetup{type=figure}
	\begin{subfigure}[b]{.18\linewidth}
		\includegraphics[width=\textwidth]{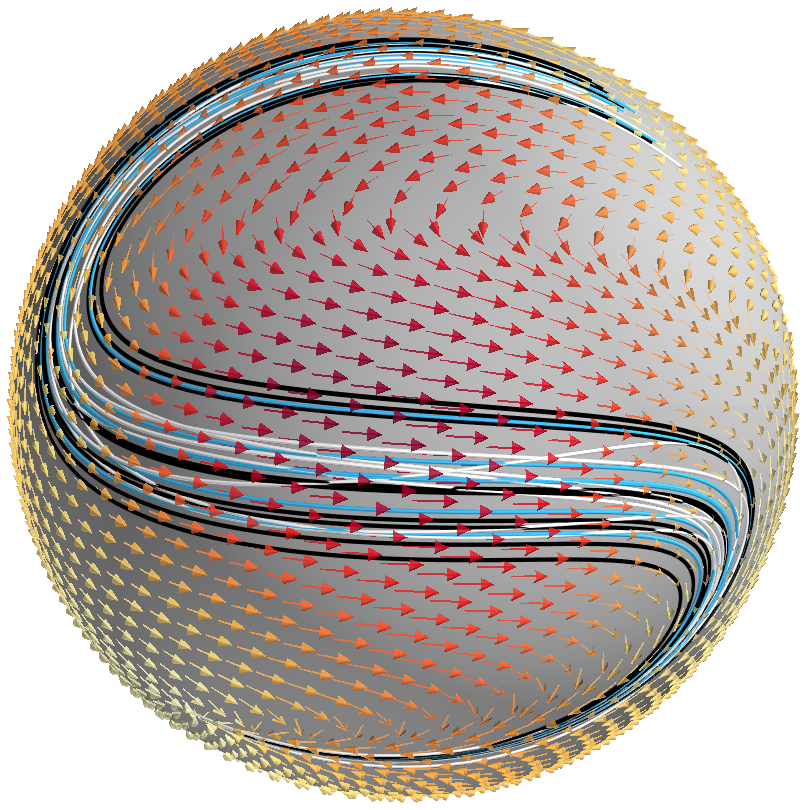}
		\includegraphics[width=\textwidth]{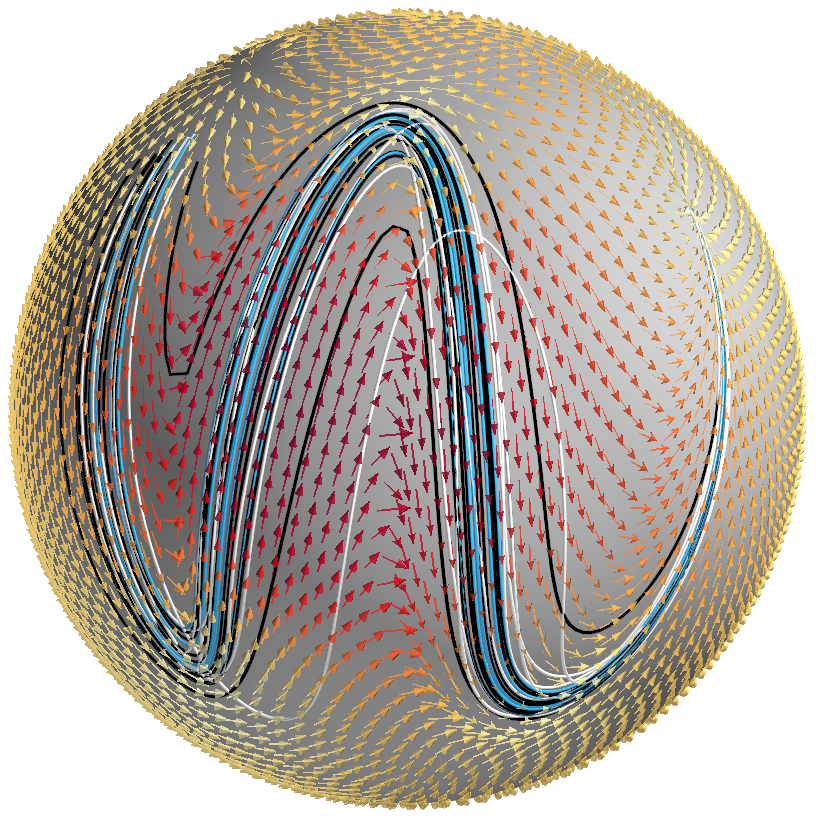}
		\caption{Riemannian DS} 
		\label{subFig:SphereRiemannianDS}
	\end{subfigure}
  \hspace{0.3cm}
	\begin{subfigure}[b]{.18\linewidth}
		\includegraphics[width=\textwidth]{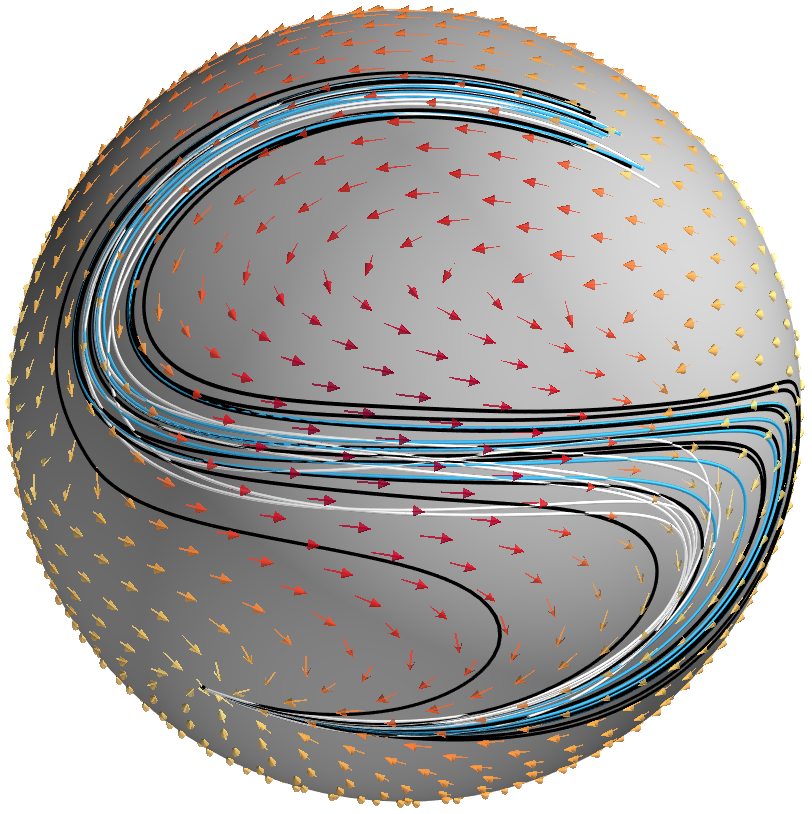}
		\includegraphics[width=\textwidth]{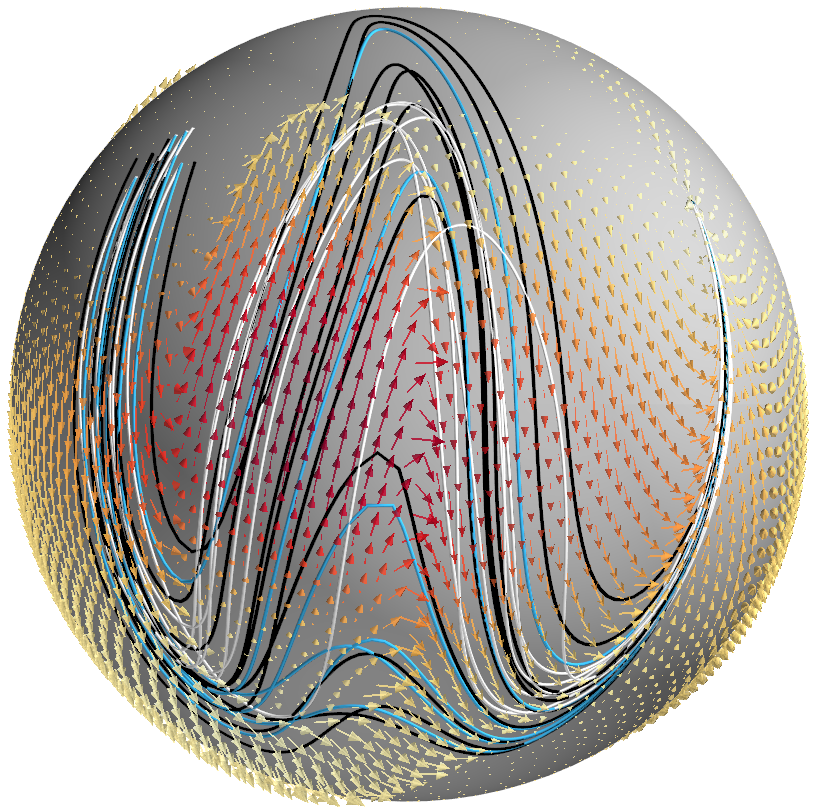}
		\caption{Projected Euclidean DS} 
		\label{subFig:SphereProjectedEuclideanDS}
	\end{subfigure}
 \hspace{0.3cm}
	\begin{subfigure}[b]{.18\linewidth}
		\includegraphics[width=\textwidth]{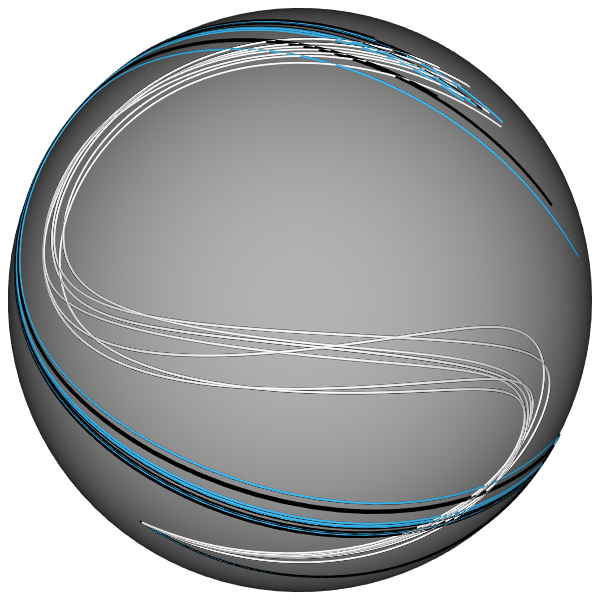}
		\includegraphics[width=\textwidth]{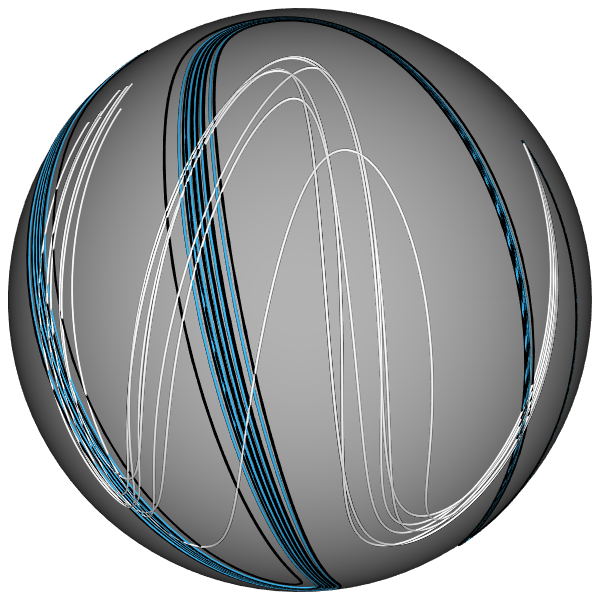}
		\caption{Tangent DS } %
		\label{subFig:SphereNaiveMeanDS}
	\end{subfigure}
 \hspace{0.3cm}
    \begin{subfigure}[b]{.185\linewidth}
		\includegraphics[width=\textwidth]{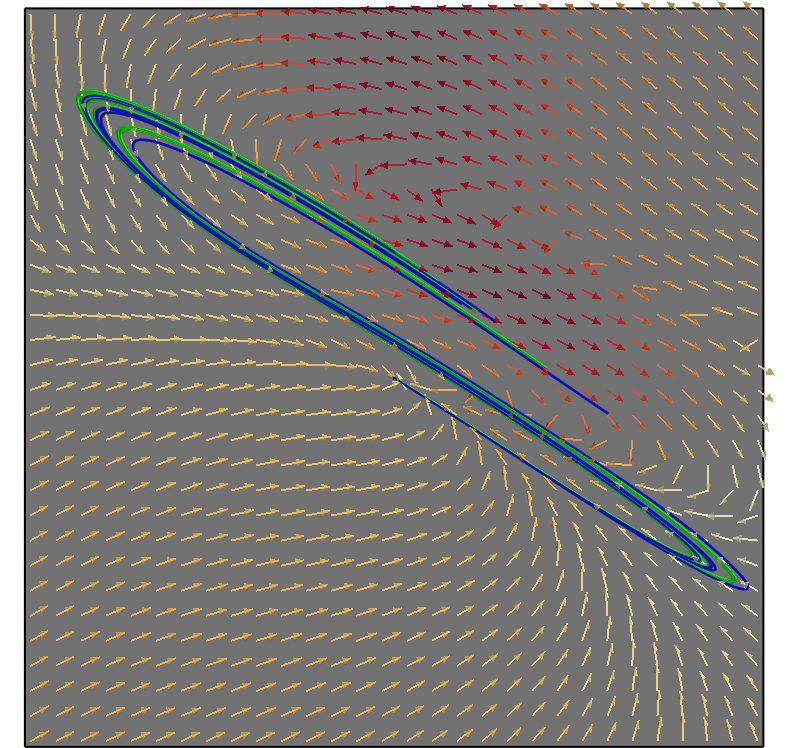}
  
        \vspace{0.2cm}
        
		\includegraphics[width=\textwidth]{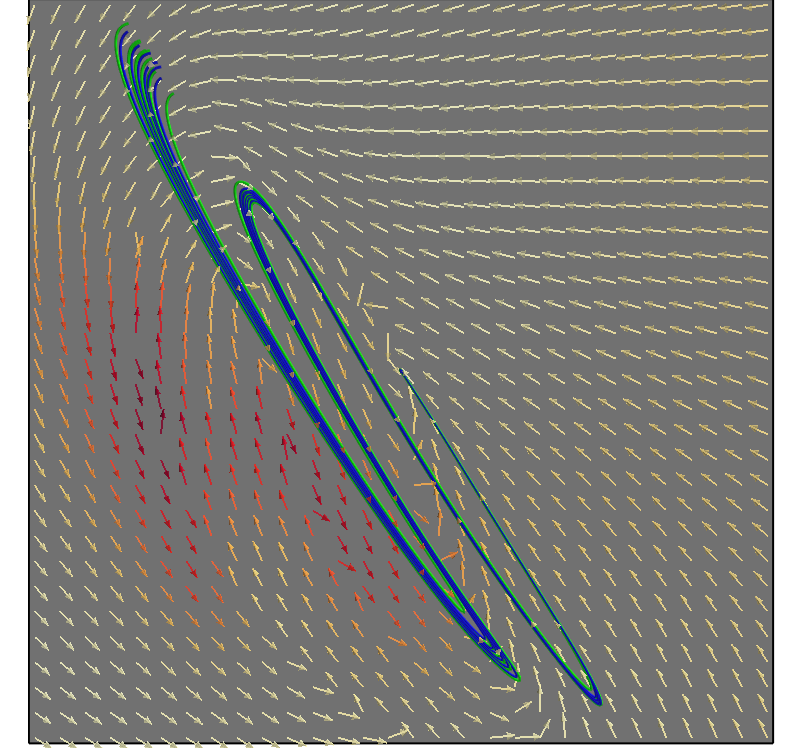}
		\caption{Tangent space projection} %
		\label{subFig:SphereNaiveMeanDS_TangentSpace}
	\end{subfigure}
	\caption{Learning first-order DS on the LASA datasets $\mathsf{S}$ \emph{(top)} and $\mathsf{W}$ \emph{(bottom)} projected on $\sphere{2}$. The demonstrations are displayed as white curves and the learned vector field is depicted by arrows (color-coded based on the magnitude). Blue trajectories are reproductions starting at the same initial points as the demonstrations, while black trajectories start from randomly-sampled points in their neighborhood. The vector field of Tangent DS is not displayed in (c) as it is learned on the single tangent space in (d) and does not produce valid velocities.}
	\label{Fig:DS_comparison}
 \vspace{-0.6cm}
\end{figure*}

\begin{table}[t]
    \setlength{\tabcolsep}{2pt}
    \renewcommand{\arraystretch}{1.05}
    \centering
    \begin{tabular}{c|cc|cc|}
        \multicolumn{1}{c|}{} & \multicolumn{2}{c|}{DTWD} &\multicolumn{2}{c|}{Success rate}\\
        Dataset & $\mathsf{S}$ &$\mathsf{W}$ & $\mathsf{S}$ & $\mathsf{W}$ \\
         \hline
        Proj. Eucl. DS &$3.2\pm3.1$ & $4.4\pm6.5$ & $0.97\pm0.02$ & $0.87\pm0.15$\\
		Tangent DS &$6.4\pm0.3$ & $9.2\pm3.0$ & $0.96\pm0.07$ & $0.67\pm0.04$\\
        Riem. DS & $\bm{0.91\pm1.2}$ & $\bm{1.1\pm3.9}$ & $\bm{1.0\pm00}$ & $\bm{0.99\pm0.02}$\\
	    \hline
 \end{tabular}
    \caption{Average DTWD w.r.t the demonstrations and success rate over $1000$ runs for Proj. Eucl. DS, Tangent DS, and Riem. DS.}
    \label{tab:ResultsRSDS}
    \vspace{-0.5cm}
\end{table}

We evaluate the effects of considering a single tangent space to learn first-order DS on Riemannian manifolds.
Given a set of demonstrations, our learning problem boils down to learn $f_{\bm{\theta}}$, which can be represented using a GMM as in~\cite{KhansariZadeh2011:StableEstimatorDS,Blocher17:LearningContractiveDS,Figueroa18:LearningPhysicsDS}, or normalizing flows~\cite{Papamakarios21:NormalizingFlows} as in~\cite{Rana2020:EuclideanizingFlows,Zhang23:RiemannianStableDS,Wang22:DeepRiemannianSkills,Urain22:LieAlgebraDS}. 
We focus on the latter as it offers a general and extensible framework to model highly-complex vector fields.
We tackle the learning problem via: \emph{(1)} a full Riemannian framework~\cite{Zhang23:RiemannianStableDS}, \emph{(2)} a projection-based Euclidean approach, projecting solutions in $\euclideanspace^d$ to $\manifold$, and \emph{(3)} a single-tangent space perspective as proposed in~\cite{Wang22:DeepRiemannianSkills}.
We evaluate all approaches on the LASA dataset~\cite{Lemme2015:LasaDataset} projected on $\sphere{2}$~\cite{Zhang23:RiemannianStableDS}, so that $\bm{x} \in \sphere{2}$, $\Dot{\bm{x}} \in \mathcal{T}_{\bm{x}}\sphere{2}$.
A dataset $\{\{\bm{x}_{m,t}, \Dot{\bm{x}}_{m,t}\}_{t=1}^{T_m}\}_{m=1}^M$ of $M$ demonstrations is used to learn $f_{\bm{\theta}}$, which is represented by a Riemannian continuous normalizing flow (CNF)~\cite{lou2020:NeuralMODE}.
The CNF is trained on a different spaces: \emph{(1)} $\sphere{2}$ for the Riemannian DS, \emph{(2)} $\euclideanspace^3$ for the projected Euclidean DS, and \emph{(3)} $\mathcal{T}_{\bm{g}}\sphere{2}$ for the Tangent DS with $\bm{g}$ being the DS attractor. 

Fig.~\ref{Fig:DS_comparison} shows the demonstrations for the datasets $\mathsf{S}$ and $\mathsf{W}$, the 
learned vector fields, and the reproduced trajectories for all considered approaches.  
As expected, the Riemannian framework provides reproductions that closely match the demonstrations. 
The projection-based approach retrieves trajectories that roughly match the demonstrations pattern, although some reproductions significantly deviate from the data support. 
The vector field learned on a single-tangent space leads to very distorted reproductions, even for starting points coinciding with those of the demonstrations. 
Despite the learned vector field on $\mathcal{T}_{\bm{g}}\sphere{2}$ capturing the demonstrations pattern, it is evident that the demonstrations are severely distorted in $\mathcal{T}_{\bm{g}}\sphere{2}$.
Such distortions are then reflected on $\sphere{2}$ when projecting the reconstructions using the exponential map. 
Note that the trajectories of both datasets cover a large part of the semisphere of $\sphere{2}$, thus being hard to approximate via a single tangent space.
Indeed, as the starting points of the demonstrations are far away from the goal, they display a high distortion which is exacerbated by the exponential and logarithmic maps needed to reproduce the motion. 

We quantitatively assess the performance of the considered approaches via \emph{(1)} the dynamic time warping distance (DTWD) as a measure of trajectory reproduction accuracy, and \emph{(2)} the stability of the learned DS as the percentage of trajectories converging to the goal out of $1000$ uniformly-sampled initial points on $\sphere{2}$.
Table~\ref{tab:ResultsRSDS} reports the DTWD and success rate metrics, where the Riemannian formulation consistently outperforms its Euclidean counterparts. 
Notice that a low success rate means that the model is unable to guarantee stable reproductions, which is imperative when controlling real robots with this kind of approaches.
\vspace{-0.05cm}

\section{TAKE HOME MESSAGES}
\vspace{-0.05cm}
This paper emphasized the critical importance of avoiding the fallacy of using a single tangent space for operating on Riemannian manifolds in robot learning problems.
We experimentally showed how the performance of machine learning methods can be severely affected by such an approximation, even in low-dimensional manifolds of constant curvature and noticeably more in complex, high-dimensional manifolds.
Good practices for operating with Riemannian manifolds in robot learning include: \emph{(1)} designing coordinate-invariant algorithms that naturally emerge from the geometric constructions defined via Riemannian metrics; \emph{(2)} leveraging a bundle of tangent spaces and parallel transport operators to capture the manifold's intricate geometry and global structure; \emph{(3)} formulating learning models that leverage the core principles of Riemannian theory. 
For example, Ijspeert et al.~\cite{Ijspeert02:OriginalDMP,Ijspeert02:PeriodicDMPs} formulated dynamic movement primitives based on second-order DS and nonlinear oscillators.
Naturally, a sound Riemannian approach should build on nonlinear second-order DS on Riemannian
manifolds~\cite{Fiori15:Riemannian2DS}, where concepts such as the covariant derivative are crucial for a proper treatment of the problem. 
Similarly, reinforcement learning (RL) algorithms on Riemannian manifolds should be formulated on a case-by-case basis by adapting their core principles. For instance, Riemannian Bayesian optimization requires Riemannian kernels and optimization techniques~\cite{Jaquier19:GaBO,Jaquier21:GaBO}. This is fundamentally opposed to using a single tangent space in which any off-the-shelf RL method is applied as in~\cite{Alhousani23:GeometricRL}.
To conclude, our main message is to embrace mathematically-sound techniques so that researchers and practitioners unlock the full potential of Riemannian manifolds in robot learning.







\clearpage
\bibliographystyle{IEEEtran}

\bibliography{References} 

\end{document}